\pdfoutput=1

\documentclass[11pt]{article}

\usepackage{acl}

\usepackage{times}
\usepackage{latexsym}

\usepackage[T1]{fontenc}

\usepackage[utf8]{inputenc}

\usepackage{microtype}

\usepackage{mathrsfs}
\usepackage{graphicx}

\usepackage{amsmath}
\usepackage{amsfonts}
\usepackage{multirow}
\hypersetup{draft}

\title{DeepE: a deep neural network for knowledge graph embedding}

\author{
    Zhu Danhao \\
    Department of Criminal Science and \\Technology, 
    Jiangsu Police Institute \\ 
     \texttt{229369897@qq.om}\\\\
    {\bf Huang Shujian}\\
    Department of Computer Science and \\Technology, 
    Nanjing University\\\\
    \And
    Shen Si \\
    School of Economics and Management,\\
    Nanjing University of Science and \\Technology \\\\
    {\bf Yin Chang}, {\bf Ding Ziqi}\\
    Department of Computer Information and \\Network Security, 
    Jiangsu Police Institute \\  
}

\begin{document}
\maketitle
\begin{abstract}
Recently, neural network based methods have shown their power in learning more expressive features on the task of knowledge graph embedding (KGE). However, the performance of deep methods often falls behind the shallow ones on simple graphs. One possible reason is that deep models are difficult to train, while shallow models might suffice for accurately representing the structure of the simple KGs.

In this paper, we propose a neural network based model, named DeepE, to address the problem, which stacks multiple building blocks to predict the tail entity based on the head entity and the relation. Each building block is an addition of a linear and a non-linear function. The stacked building blocks are equivalent to a group of learning functions with different non-linear depth. Hence, DeepE allows deep functions to learn deep features, and shallow functions to learn shallow features. Through extensive experiments, we find DeepE outperforms other state-of-the-art baseline methods. A major advantage of DeepE is the robustness. DeepE achieves a Mean Rank (MR) score that is 6\%, 30\%, 65\% lower than the best baseline methods on FB15k-237, WN18RR and YAGO3-10. Our design makes it possible to train much deeper networks on KGE, e.g. 40 layers on FB15k-237, and without scarifying  precision on simple relations.\footnote{The code and data of DeepE will be released around 2022.11.30. https://github.com/zhudanhao/DeepE}
\end{abstract}

\section{Introduction}
Knowledge graphs (KGs) are collections of facts. Some well-known knowledge graphs include Freebase~\cite{bollacker2008freebase}, WordNet~\cite{miller1995wordnet},YAGO~\cite{suchanek2007yago} and NELL~\cite{mitchell2018never}, are proved to be effective for a variety of downstream applications, such as question answering~\cite{ferrucci2010building}, information extraction~\cite{mintz2009distant} and recommender systems~\cite{zhang2016collaborative}.

Real world KGs suffer from the problem of incompleteness. The task of Knowledge Graph Embedding (KGE) learns low-dimensional feature vectors for entities and relations, and then predicts the missing facts. The core of a KGE method is the defined score function. The score of a valid fact is expected to be higher than a invalid fact. One main category of KGE methods is built based on shallow score functions, e.g., TranE~\cite{bordes2013translating}, Distmult~\cite{lin2015learning}, Rescal~\cite{nickel2011three} and RotatE~\cite{sun2019rotate}. More recently, neural network based models are proposed to learn more expressive features. The representative methods include ConvE~\cite{dettmers2018convolutional} and the following works, such as HypER~\cite{balavzevic2019hypernetwork}, InteractE~\cite{vashishth2020interacte} and JointE~\cite{zhou2022jointe}.

However, deep models do not always result in precision improvement. ConvE\cite{dettmers2018convolutional} is found to have an advantage over shallow models on complex graphs that contain nodes with high average relation-specific indegree. But on simple graphs with low average indegree, ConvE cannot outperform its linear baseline methods. Similar phenomenon has also been observed by \citet{vashishth2020interacte}.  \citet{dettmers2018convolutional}'s explanation is that deep models are difficult to train, while shallow models  might suffice for accurately representing the structure of the simple KGs.

Unlike the task of image recognition or language processing, one distinct characteristic of KGE is that shallow features also matter a lot. For example, the pix level features can hardly serve for identifying a dog directly. In contrast, a large proportion of relations in knowledge graphs are linear, such as symmetry/antisymmetry, inversion and composition~\cite{sun2019rotate}. Shallow functions are enough for scoring these linear patterns. Although deep functions are more expressive in theory, they may not learn as good as shallow functions on simple relations, whether because of overfitting or training difficulty.

Is there anyway to build a  model that can enjoy the ability of learning deep features, but without paying the price of losing shallow features? In this paper, we propose a novel deep neural network, named DeepE, to achieve the goal. The key component of DeepE is the DeepE building block, whose output is an addition of both linear and non-linear features. Stacking $n$ building blocks will obtain a learning function consists $n+1$ sub-functions, with their non-linear depth range from $0$ to $n$. Then, each sub-function is expected to be responsible for its own duty: deep function for learning deep feature, and shallow function for learning shallow feature.

The architecture of the hybrid learning functions has two consequences. First, now it is possible to train very deep networks without losing precision on simple relations. For example, on FB15-237k, the best DeepE model contains 40 layers, while previous methods can afford  1-4 layers only. Second, DeepE is robust in various situations, especially when data is sparse or relations are difficult. The result will not be too bad as long as one sub-function works.

The contributions of the paper are summarized as follows.
\begin{itemize}
  \item We propose DeepE, a KGE method that can learn very deep features, without sacrificing the ability of learning shallow features.
  \item We provide theoretical analysis to the key component of our method, the DeepE building block, to show why DeepE can learn features with different non-linear depth.
  \item Through extensive experiments on various simple and complex KG datasets, we demonstrate the effectiveness of DeepE.
\end{itemize}

\section{Definitions}
\subsection{Problem definition}
A knowledge graph $\mathcal{G} = \{(h,r,t)\} \subseteq \mathcal{E} \times \mathcal{R} \times \mathcal{E}$ is formalized as a set of knowledge triples or facts, each consists of a relation type $r \in \mathcal{R}$ points from a head entity $h \in \mathcal{E}$ to a tail entity $t \in \mathcal{E}$.

Predicting the missing links of KG can be formalized as a ranking problem. A KGE model first maps the entities and relations to vectors $\textbf h$, $\textbf r$, $\textbf t$, and then defines a score function $\Psi(\textbf h, \textbf r, \textbf t)$ that maps the triple to a scalar which is proportional to the likelihood of the triple. To predict the tail entity $t$ for $(h,r,?)$, $t = \mathop{\arg\max}\limits_{t \in \mathcal{E}} \Psi(\textbf h$, $\textbf r$, $\textbf t)$. Similar process can be used for finding a head entity $(?,r,t)$. In this paper, we do not distinguish head prediction with tail prediction. For a triple $(h,r,t)$ in the training set, we will insert an reverse triple $(t,r',h)$ where $r'$ is the reverse relation of $r$. On test stage, finding $(?,r,t)$ is equivalent to solving $(t,r',?)$.

\subsection{Non-linear depth of functions}

\newtheorem{Definition}{Definition}[section]

We formally define of the non-linear depth of a function. A deeper function contains more number of nested non-linear functions.

\begin{Definition}[0th order non-linear function]
A 0th order non-linear function is a function with no non-linear transformations.
\end{Definition}

\begin{Definition}[kth order non-linear function]
A kth order non-linear function is a non-linear transformation of a k-1th order non-linear function.

The linear combination of a kth and a jth (j<=k) order non-linear function is a kth order non-linear function.
\end{Definition}

The linear methods belong to the family of 0th order non-linear functions, and most of the existing neural network based methods are 1th to 4th non-linear functions. The higher order of a non-linear function, the deeper it is, and the deeper features are produced.

\section{The proposed method}

\subsection{Overall framework}
The overall framework of DeepE is shown in Fig. 1. DeepE is composed of two multi-layer neural networks, named feature extraction network and project network. The formal network learns features from the head entity and the relation, while the latter one projects the tail entity to the same space of the learned features. Similar to the previous works~\cite{dettmers2018convolutional,vashishth2020interacte}, $h,r,t$ are mapped to embeddings  $\textbf h,\textbf r,\textbf t \in \mathbb R^d$ first, $d$ denotes the dimension of embeddings.

\begin{figure}[tb]
\label{fig:1}
\begin{center}
{\includegraphics[width=1\linewidth]{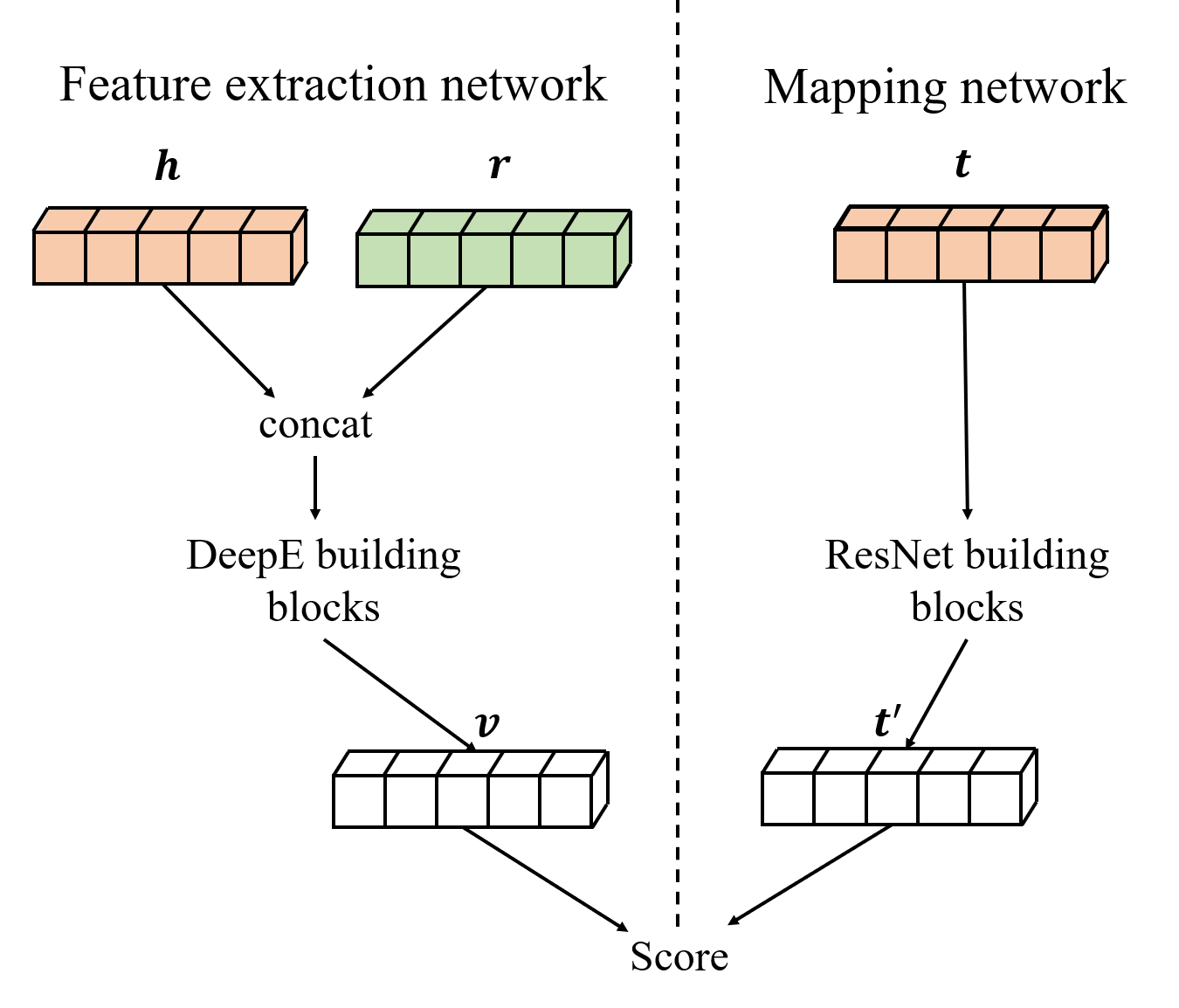}}
\caption{An overview of DeepE.}
\end{center}
\end{figure}

\subsection{Feature extraction network}
The feature extraction network first concatenates $\textbf h$ and $\textbf r$, and then let it go through a BN layer and a dropout layer. The obtained vector is used as an input for multiple stacked DeepE building blocks. The building blocks return a feature vector $\textbf v \in \mathbb R^d$. 

DeepE building block is the basic component of feature extraction network, and we use stacked DeepE building blocks to extract features from head entity and relation.

\subsubsection{A DeepE building block}

\begin{figure}[tb]
\label{fig:1}
\begin{center}
{\includegraphics[width=1\linewidth,height=4.5cm]{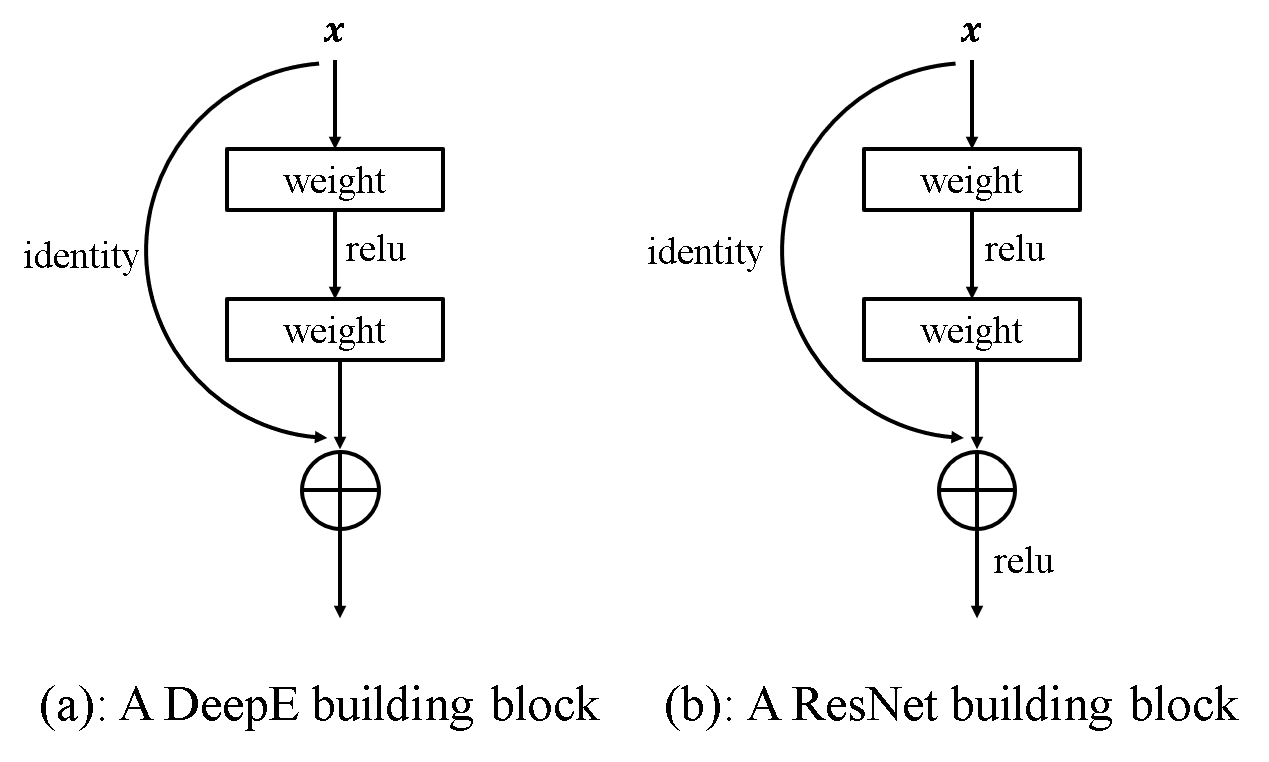}}
\caption{The building blocks. We omit the BN and Dropout for concise.}
\end{center}
\end{figure}
The structure of a DeepE building block is shown in Fig. 1 (a). $\textbf{x}$ and $\mathcal F(\textbf x)$ are the input and output vectors of a building block. Formally, a DeepE building block is defined as follows:

\begin{equation}
    \mathcal F(\textbf{x}) = \textbf{x} + \textbf W_2\sigma(\textbf W_1\textbf{x})
\end{equation}

$\textbf W_1$ and $\textbf W_2$ are the weight matrices. $\sigma$ is the non-linear function, and we use ${\rm Relu}$ across the paper. For convenience, we omit the bias terms, dropout layer and Batch Normalization (BN) layer here. Generally, a BN layer~\cite{ioffe2015batch} and a dropout layer~\cite{srivastava2014dropout} are inserted right after each linear layer, to reduce overfitting and stabilize the training.

The proposed DeepE building block is quite similar to the one of ResNet~\cite{he2016deep}, as shown in Fig. 1 (b). The only difference is that DeepE building block removes the final non-linear layer of ResNet building block. Hence, a DeepE building block outputs a linear feature vector $\textbf x$,  and a non-linear feature vector $\textbf W_2\sigma(\textbf W_1\textbf{x})$. In contrast, a ResNet building block can only produce a non-linear feature vector $ \sigma(\textbf{x} + \textbf W_2\sigma(\textbf W_1\textbf{x}))$.

If the dimensions of $\textbf x$ and $\textbf W_2\sigma(\textbf W_1\textbf{x})$ are not equal, a linear projection $\textbf W_s$ will be applied to $\textbf x$, to make the element-wise addition available.
\begin{equation}
    \mathcal F(\textbf{x}) = \textbf W_s \textbf{x} + \textbf W_2\sigma(\textbf W_1\textbf{x})
\end{equation}

In the feature extraction network, the input and output dimensions of the first building block are $2d$ and $d$ respectively, as in Eqn. 2, where $\textbf W_s \in \mathbb R^{d*2d}$. The other building blocks share the same input and output dimensions of $d$, so the compute function is as in Eqn. 1.

\subsubsection{Stacking multiple building blocks}

The identity mapping of $\textbf x$ can well address the problem of network degradation~\cite{he2016deep}, which allows training a network with multiple stacked building blocks. Considering a simple case of stacking 2 building blocks. The superscript $(i)$ denotes the ith building block.
\begin{equation}
\begin{aligned}
    \mathcal F^{(2)} (\mathcal F^{(1)}(\textbf{x}))=   \textbf x  + \textbf W_2^{(1)}\sigma(\textbf W_1^{(1)}\textbf x)+ \\ \textbf W_2^{(2)}\sigma(\textbf W_1^{(2)}(\textbf x  + \textbf W_2^{(1)}\sigma(\textbf W_1^{(1)}\textbf x) )
\end{aligned}
\end{equation}

Eqn.3 contains 3 terms, whose non-linearity orders are $0,1,2$ respectively. It can be easily derived that when stacking $n$ DeepE building blocks, the output will consist $n+1$ terms, and their orders of non-linearity should be $0,1,2..n$ respectively. Therefore, deep and shallow features can be easily learned by functions with different non-linear depth. In contrast, stacking $n$ ResNet building blocks can only produce a non-linear function with $2n$th depth, which is not suitable for extracting shallow features. Our experiments compare the performance of ResNet and DeepE building blocks in section 4.3.2.

It is not recommended to add more non-linear layers in a DeepE building block, i.e. $ \mathcal F(\textbf{x})  = \textbf{x} + \textbf W_n \sigma (\textbf W_{n-1}... \sigma(\textbf W_2\sigma(\textbf W_1\textbf{x}))...)$. Considering a case of stacking $n$ building blocks, each with 2 inner non-linear layers. The output will consist $n+1$ terms with non-linear orders of $0,2,4,..2n$. Clearly, the more inner non-linear layers a building block contains, the more levels of non-linear order the stacked architecture will loss.

\subsubsection{Dropout on identity mapping}
Although identity mapping is benefit for training deep network, it may cause the problem of diminishing feature reuse~\cite{srivastava2015highway}. The gradients will prefer to go through the identity mappings rather than the weights within building blocks.

To address the problem, we propose to dropout the identity mapping with a small ratio $\alpha$. For a network of $n$ stacked building blocks, the $i$th order of non-linear feature vector has to go through the identity mappings for $n-i$ times. Hence, its total dropout probability will be $(1-\alpha)^{n-i}$. Such design will gradually dropout more shallow features than deep features. For example, for a DeepE model with 40 building blocks (the one we use for FB15k-237), $\alpha=0.01$, the final dropout probability of the $0$th, $10$th, $20$th and $30$th order of non-linear features will be 0.331, 0.260, 0.182, 0.096.

Note that \citet{he2016identity} also experimented dropout on the identity mapping in ResNet, and found the network failed to converge to a good solution. One probable reason is that they used a dropout value of 0.5, which is too big. As discussed before, such a large value will impede signal propagation.

\subsection{Project network}
$\textbf v$ is obtained via a series of functions with different orders of non-linearity, which is in a space quite different from the original entity space. Therefor, it will be better to project $\textbf t$ to a space close with $\textbf v$. $\textbf t$ is used as an input for a stacked ResNet building blocks, and returns a vector $\textbf t'$. In practice, the number of ResNet building blocks are no more than 2.

\subsection{Score function}
The score function is the dotted product between $\textbf  v$ and $\textbf  t'$. Overall,  our score function is written as follows:

\begin{equation}
\begin{split}
    \Psi(h,r,t) &= {\rm f}(\textbf h || \textbf r) \cdot {\rm g}(\textbf t)
    \\
    &= \textbf  v  \cdot \textbf t'
\end{split}
\end{equation}

where $||$ is the concatenation operator, ${\rm f}$ and ${\rm g}$ are the feature extraction network and project network respectively. For training, we use standard cross entropy loss function.

\subsection{Space complexity}
There are three types of parameters in DeepE. First, the embedding matrices for entities and relations, with size of $|\mathcal{E}|d+|\mathcal{R}|d$. Second, the parameters of the DeepE building blocks in the feature extraction network. Each building block has two matrices, and the parameter size is  $2kd^2$, where $k$ is the number of DeepE building blocks. Third, the parameters of the ResNet building blocks in the project network whose parameter number is about $2td^2$, $t$ denotes the number of ResNet building blocks.

In general, $|\mathcal{R}|$ and $d$ are much smaller than $|\mathcal{E}|$, and $t$ is no more than 2. The final space complexity can be reduced to:
$$
    \mathcal{O}(|\mathcal{E}|d + kd^2)
$$
On large KGs with a massive mount of entities, DeepE is very parameter efficiency since $kd^2$ can also be neglected.

\section{Experiments}
In this section, we want to investigate the following research questions:

RQ1. How does DeepE perform in compare with the baseline methods? (Section 4.2)

RQ2. What is the impact of different modules in DeepE? (Section 4.3)

RQ3. What is the impact of functions with different order of non-linearity? (Section 4.4)

RQ4. Why DeepE is robust than other methods? (Section 4.5)

\subsection{Experiment setup}
\subsubsection{Datasets}
Following ~\cite{dettmers2018convolutional,sun2019rotate,vashishth2020interacte}, three most common benchmark datasets are used. The summary statistics of the datasets are presented in Table 1.

\begin{table}
\centering
\footnotesize{
\begin{tabular}{cccc}
\hline
Dataset&FB15k-237&WN18RR&YAGO3-10 \\
$|\mathcal{E}|$ & 14541 & 40943 & 123182\\
$|\mathcal{R}|$ & 237 & 11 & 37\\
\#Train & 272115 & 86835 & 1079040\\
\#Valid & 17535 & 3034 & 5000\\
\#Test & 20446 & 3134 & 5000\\
\hline
\end{tabular}
}
\caption{Statistics of the datasets. \# denotes the number of  triples.}
\label{tab:accents}
\end{table}

\begin{itemize}
  \item FB15k-237~\cite{toutanova2015observed} is a improvement version of FB15k~\cite{bordes2013translating} dataset where the some inverse relations are removed, to avoid data leakage.
  \item WN18RR~\cite{dettmers2018convolutional} is a subset of WN18~\cite{bordes2013translating}, where the inverse relations are also removed.
  \item YAGO3-10~\cite{suchanek2007yago} is a subset of YAGO3. Most of the triples deal with descriptive attributes of people.
\end{itemize}

\subsubsection{Evaluation protocol}
Four metrics are used to measure the performance, including Mean Rank (MR), Mean Reciprocal Rank (MRR), Hit@1 and Hit@10. We follow the filtered settings in \citet{bordes2013translating}, that is, excluding all true entities appear in the train, valid and test sets when ranking all entities. Note that models with higher MRR, Hit@1, Hit@10 and lower MR are preferred.

\subsubsection{Baseline methods}
We compare our method with various KGE methods, which can be classified into three categories.
\begin{itemize}
  \item Shallow methods, whose score functions contain no non-linear functions, include DistMult~\cite{yang2014embedding} and RotatE~\cite{sun2019rotate}.
  \item Convolutional neural network (CNN) based methods. These CNN based methods are the dominating techniques among non-linear methods,  include ConvE\cite{dettmers2018convolutional}, HypER~\cite{balavzevic2019hypernetwork}, InteractE~\cite{vashishth2020interacte}, AcrE~\cite{ren2020knowledge} and JointE~\cite{zhou2022jointe}.
  \item  A multi-layer perception (MLP) method, ER-MLP-2d~\cite{ravishankar2017revisiting}. Although ER-MLP-2d is not a very strong baseline, both the method and DeepE are based on MLP, rather than CNN. Hence, our method is closer to ER-MLP-2d than other baseline methods.
\end{itemize}

\subsubsection{Training details}
All parameters are random initialized with xavier normal distribution~\cite{glorot2010understanding}. We use Adam~\cite{kingma2014adam} optimizer with an initial learning rate of 0.003. The learning rate will decrease with a coefficient of 0.8 when the training loss does not decrease for 5 epoches. The other training parameters for each dataset are shown in Table 2. When the MRR does not improve for 10 epochs on the valid set, training will be terminated. The best models are selected when the best MRR are obtained on the valid set. We report the average results of 5 runs with different randomly initialization. The maximum training epoch is 1000. In practice, WN18RR and YAGO will finish training in about 200 epochs, and FB15k-237 takes about 700 epochs.

\begin{table}
\centering
\footnotesize{
\begin{tabular}{cccc}
\hline
&FB15k-237&WN18RR&YAGO3-10 \\

P1 & 300& 250 &500 \\
P2 & 5e-8& 5e-5 & 5e-8\\
P3 & 40& 1 & 2\\
P4 & 1& 2 & 1\\
P5 & 2& 3 & 2\\
P6 & 0.4& 0.4&0.4\\
P7 & 0.01& 0&0\\
P8 & 0.4& 0& 0\\
\hline
\end{tabular}
}
\caption{Parameter settings for each dataset. P1: Entity/Relation dimension. P2: L2 regularization. P3: Number of DeepE building blocks. P4: Number of ResNet building blocks. P5: Number of inner layers in ResNet building blocks. P6: Dropout on input layer and FC layers in DeepE building blocks. P7: Dropout on the identity mapping in DeepE building blocks. P8: Dropout on the FC layers in ResNet building blocks. }
\label{tab:accents}
\end{table}

\subsection{Performance Comparison}
\subsubsection{Main results}
\begin{table*}
\centering
\footnotesize{
\begin{tabular}{lllll|llll|llll}
    \hline
        \multirow{1}*{Model} &  \multicolumn{4}{c|}{FB15k-237}  &  \multicolumn{4}{c|}{WN18RR}  &  \multicolumn{4}{c}{YAGO3-10}  \\
         & MR & MRR & Hit@1 & Hit@10 & MR & MRR & Hit@1 & Hit@10 & MR & MRR & Hit@1 & Hit@10 \\
        \hline
        Distmult & 254 & 0.241 & 0.155 & 0.419 & 5110 & 0.43 & 0.39 & 0.49 & 5926 & 0.34 & 0.24 & 0.54  \\
        RotatE & 177 & 0.338 & 0.241 & 0.533 & \underline{3340} & \underline{0.476} & 0.428 & \textbf{0.571} & 1767 & 0.495 & 0.402 & 0.67  \\ \hline
        ConvE & 244 & 0.325 & 0.237 & 0.501 & 4187 & 0.43 & 0.4 & 0.52 & \underline{1671} & 0.44 & 0.35 & 0.62  \\
        HypER & 250 & 0.341 & 0.252 & 0.52 & 5798 & 0.465 & 0.436 & 0.522 & 2529 & 0.533 & 0.455 & 0.678  \\
        InteractE & \underline{172} & 0.354 & 0.263 & 0.535 & 5202 & 0.463 & 0.43 & 0.528 & 2375 & 0.541 & 0.462 & 0.687  \\
        AcrE & - &\textbf{0.358} & \textbf{0.266} & \underline{0.545} &- &0.459 &0.422 &0.532 &- & -&- & -\\
        JointE & 177 & 0.356 & 0.262 & 0.543 & 4655 & 0.471 & \underline{0.438} & 0.537 & - & \underline{0.556} & \underline{0.481} & \textbf{0.695}  \\ \hline
        ER-MLP\-2d & 234 & 0.338 & - & \textbf{0.547} & 4233 & 0.358 & - & 0.421 & - & - & - & -   \\ \hline
        DeepE & \textbf{161} & \textbf{0.358} & \textbf{0.266} & 0.544 & \textbf{2337} & \textbf{0.487} & \textbf{0.445} & \underline{0.567} & \textbf{591} & \textbf{0.558} & \textbf{0.485} & \underline{0.692} \\
        \hline
    \end{tabular}
}
\caption{The main results. The best results are in bold, the second best results are underlined. }
\end{table*}

The main results are presented in Table 3. Overall, DeepE achieves the best and the second best results on most metrics. In particular, our method outperforms the baseline methods on MR with a large margin, i.e. 172 -> 161 (-6\%) on FB15k-237, 3340->2337 (-30\%) on WN18RR, and 1671->591 (-65\%) on YAGO3-10. Comparing to MRR, Hit@1 and Hit@10, MR is a metric more concerning about robustness, since a bad prediction will pull down the value a lot. We will analyze the robustness of DeepE in section 4.5.

FB15k-237 is a dataset suitable for deep methods mostly. On FB15k-237, InteractE, AcrE and JointE outperform RotatE on almost all the metrics. On YAGO3-10, the advantage is less significant. InteractE and JointE are better than RotatE on 3 out of 4 metrics. But on WN18RR, the situation changes. RotatE outperforms all other non-linear methods on MR, MRR and Hit@10. The parameter settings of our method (Table 2) also validate the phenomenon. The number of building blocks for the best DeepE models are 40, 2 and 1 on FB15k-237, YAGO3-10 and WN18RR. However, unlike other non-linear methods that suffer from the problem of inconsistent performance, DeepE can work well on all three datasets.

As a closer baseline that based on MLP, ER-MLP2d's performance is much falling behind DeepE. Note that DeepE maybe the first MLP based method that achieves SOTA results, which validates the effectiveness of our motivation.

\subsubsection{Additional comparison}
Some recent studies try to exploit graph structure information for link prediction on knowledge graphs. These methods often use KGE methods as tools for extracting features~\cite{dai2022mrgat,wu2021disenkgat}. Since the score function of a graph based method is built on a sub-graph rather than on a single triple, it is meaningless  to compare the non-linear orders between these methods and DeepE. Table 4 presents the results of some SOTA graph based methods as an additional comparison with DeepE, include MRGAT~\cite{dai2022mrgat}, CompGCN~\cite{vashishth2019composition}, ReinceptionE~\cite{xie2020reinceptione} and DisenKGAT~\cite{wu2021disenkgat}. Although DeepE does not model graph structure information explicitly, it can still achieve first or second best results on most metrics.

\begin{table}
\centering
\footnotesize{
\begin{tabular}{l|ll|ll}
    \hline
        \multirow{1}*{Model} &  \multicolumn{2}{c|}{FB15k-237}  &  \multicolumn{2}{c}{WN18RR}    \\
         & MR & MRR  & MR & MRR    \\
        \hline
         MRGAT & - & 0.355 & - & 0.481 \\
        CompGCN & 197 & 0.355 & 3533 & 0.479 \\
        ReinceptionE & \underline{173} & 0.349 & \underline{1894} & 0.483 \\
        DisenKGAT & 179 & \textbf{0.368} & \textbf{1504} & \underline{0.486} \\ \hline
        DeepE & \textbf{161} & \underline{0.358}  & 2337 & \textbf{0.487}  \\
        \hline
    \end{tabular}
}
\caption{An additional comparison between DeepE and some graph based methods.}
\end{table}

\subsection{Effect of different modules}
\subsubsection{Ablation analysis}

The result of ablation analysis is shown in Table 5. Project network is essential for the performance. Without the project network, MRR decreases 0.016 on FB15k-237, and 0.026 on WN18RR. Without the identity dropout, MRR decreases 0.003 on FB15k-237. As discussed in section 2.4, identity dropout is designed for training very deep networks. We do not use it in WN18RR and YAGO3-10, since their best models only contain 1 or 2 DeepE building blocks.

\begin{table}
\centering
\footnotesize{
    \centering
    \begin{tabular}{l|ll}
    \hline
        Model & FB15k-237 & WN18RR \\
        \hline
        Original & 0.358& 0.487 \\
        - project network & 0.342 (-0.016) & 0.461 (-0.026) \\
        - identity dropout & 0.355 (-0.003) & - \\
        \hline
    \end{tabular}}
\caption{Ablation analysis. MRR is used as the metric. \emph{- mapping network} replaces the project network with an identity mapping. \emph{- identity dropout} removes the dropout layer on identity mapping. }
\end{table}

\subsubsection{Effect of depth}

\begin{figure}[tb]
\label{fig:5}
\begin{center}
{\includegraphics[width=1\linewidth]{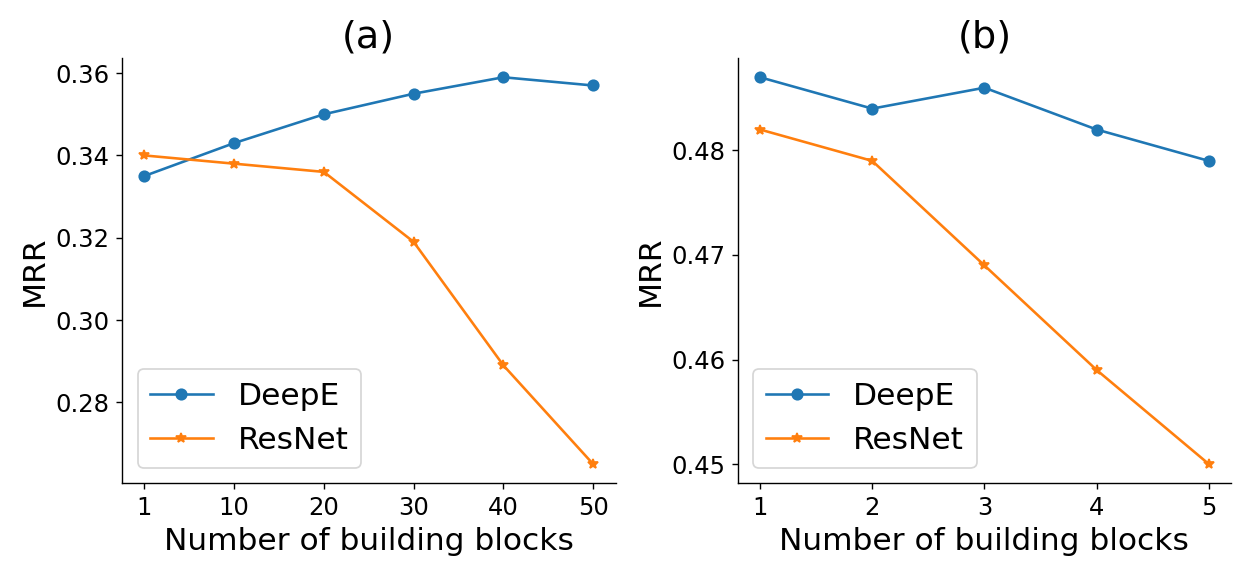}}
\caption{The effect of depth. (a) On FB15k-237. (b) On WN18RR. DeepE is our method, while ResNet denotes that we replace DeepE building blocks with ResNet building blocks in the feature extraction network. }
\end{center}
\end{figure}
Fig 3 gives the performance on different model depth, i.e. number of building blocks. On FB15k-237, the MRR of the ResNet building blocks drops quickly as the network going deeper. In contrast, DeepE gains precision improvement with deeper layers. The result shows that DeepE building block is essential for training deeper networks for KGE.

On WN18RR, both DeepE or ResNet buidling blocks perform worse when adding more layer, but the result of DeepE is more robust. When stacking a new building block, a network based on ResNet building blocks will replace the old learning function with a new one, while DeepE adds a new function to the original learning function without modifying it. Even if the new function is not suitable, DeepE's performance will not decrease too much, since the original learning function still works. We believe this is why the building block of DeepE is more robust than that of ResNet on different depth.

\subsection{Effect of functions with different non-linear orders}
\subsubsection{A single DeepE building block}
A model with only one DeepE building block is trained on FB15k-237, to investigate the effect of the linear and non-linear functions of a building block. The results are shown in Table 6. Th 0th order non-linear function suffices for the accurate prediction on simple relations of N-1.  The main contribution of the 1th order non-linear function is on the complex relations of 1-N (with an improvement of 59\% on MRR).

The results validate our hypothesis. For a single DeepE building block, the linear function is responsible for learning simple relations, and the non-linear function mainly serves for complex relations.

\begin{table}
\centering
\footnotesize{
    \centering
    \begin{tabular}{l|ll}
    \hline
        Rel type& 0th & 0th+1th    \\         \hline
        1-1 & 0.287  & 0.337 (+17\%) \\
        N-1 & 0.679  &0.701 (+3\%) \\
        1-N & 0.064  & 0.102 (+59\%) \\
        N-N & 0.27  & 0.31 (+15\%)  \\ \hline
    \end{tabular}}
\caption{The effect of  functions in a DeepE building block. MRR is used as the metric. 0th indicates that the function with non-linear order of 0th is retained, while the other function is disabled.  Following \citet{bordes2013translating}, relations are classified into four groups: one-to-one (1-1), one-to-many (1-N), many-to-one (N-1) and many-to-many (N-N). For brevity, we do not distinct head prediction with tail prediction. The tail prediction of 1-N relations is equivalent to the head prediction of N-1 relations, and vice verse.}
\end{table}

\subsubsection{Stacked DeepE building blocks}
\begin{figure}[tb]
\label{fig:5}
\begin{center}
{\includegraphics[width=1\linewidth]{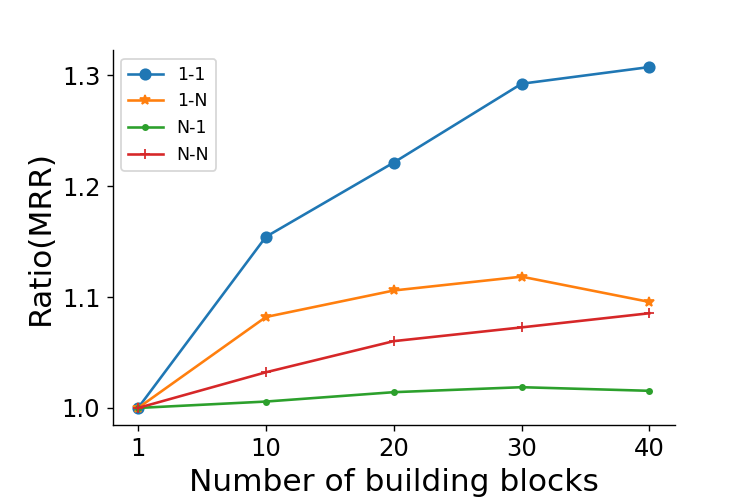}}
\caption{The effect of functions by stacked DeepE building blocks. To better reveal the tendency, the y-axis is the ratio to the model with one DeepE building block.}
\end{center}
\end{figure}

We train several models with different number of DeepE building blocks. More building blocks result in more orders of non-linear learning functions. The results are shown in Fig. 4. DeepE has addressed the problem of training deep neural networks on KGE: deeper non-linear functions no longer degrade the learning on simple relations. Relations in all groups achieve consistent improvement. For the simplest relation group, N-1, the curve is almost steady on all layers. The results validate the advantage of using different order of non-linear learning functions on KGE.

\subsection{Robustness}

\begin{figure}[tb]
\label{fig:5}
\begin{center}
{\includegraphics[width=1\linewidth]{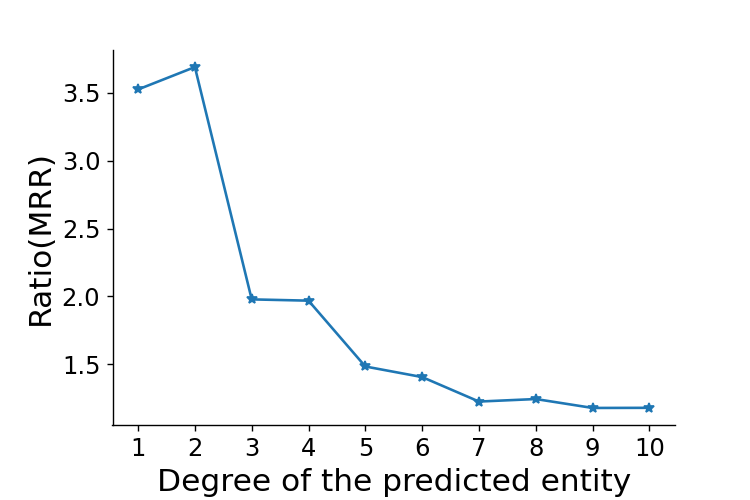}}
\caption{Performance comparison on entities with low degrees. The y-axis is the ratio of DeepE to InteractE. The results of InteractE are calculated by their released model.}
\end{center}
\end{figure}

\begin{table*}
\centering
\footnotesize{
    \centering
    \begin{tabular}{ll|lll|lll|lll|lll}
    \hline
    & & \multicolumn{3}{c|}{RotatE}  & \multicolumn{3}{c|}{ConvE} & \multicolumn{3}{c|}{InteractE} & \multicolumn{3}{c}{DeepE} \\
    & & MRR & MR & Hit@10 & MRR & MR & Hit@10 & MRR & MR & Hit@10 & MRR & MR & Hit@10 \\ \hline
    \multirow{4}*{\rotatebox{90}{Head Pred}} & 1-1 & \textbf{0.498} & 359 & 0.593 & 0.374 & 223 & 0.505 & 0.386 & \textbf{175} & 0.547 &0.458 &548 &\textbf{0.615} \\
    &1-N & 0.092 & 614 & 0.174 & 0.091 & 700 & 0.17 & 0.106 & 573 & 0.192 & \textbf{0.121} & \textbf{519} & \textbf{0.222} \\
    &N-1 & \textbf{0.471} & 108 & \textbf{0.674} & 0.444 & 73 & 0.644 & 0.466 & \textbf{69} & 0.647 & 0.461& 102&0.657 \\
    &N-N & 0.261 & 141 & 0.476 & 0.261 & 158 & 0.459 & 0.276 & 148 & 0.476 & \textbf{0.282} & \textbf{139}  & \textbf{0.485} \\
    \hline
    \multirow{4}*{\rotatebox{90}{Tail Pred}} & 1-1 & \textbf{0.484} & 307 & \textbf{0.578} & 0.366 & \textbf{261} & 0.51 & 0.368 & 308 & 0.547 &0.424 &358 &0.573 \\
    &1-N & 0.749 & 41 & 0.674 & 0.762 & 33 & 0.878 & 0.777 & \textbf{27} & 0.881 & \textbf{0.789}&47 & \textbf{0.886} \\
    &N-1 & 0.074 & 578 & 0.138 & 0.069 & 682 & 0.15 & 0.074 & 625 & 0.141 & \textbf{0.08}&\textbf{576} &\textbf{0.162} \\
    &N-N & 0.364 & 90 & 0.608 & 0.375 & 100 & 0.603 & \textbf{0.395} & 92 & \textbf{0.617} & 0.39&\textbf{83} &0.616 \\
    \hline
    \end{tabular}}
\caption{Link prediction results by relation category on FB15k-237.}
\end{table*}
In this subsection, we show DeepE is robust on different entities and relations, especially when data is sparse and learning is difficult. Since DeepE is indeed a hybrid  learning functions with different non-linear depth, the prediction result will not be too bad as long as one function works. In contrast, if a learning case relies on deep features, a method based on shallow functions may fail, and vice verse. The problem will be more severe when data is sparse.
\subsubsection{Performance on entities with low degrees}
Learning on  entities with low degree is difficult, since the number of training samples are small. However, these entities make up a large proportion of the total entities. For example, 25\% entities in FB15k-237 has a degree smaller than 10.

We compare the performance of low-degree entities between DeepE and InteractE on FB15k-237, as in Fig. 5. DeepE is more robust on low degree entities. Specially, the MRR of DeepE is about 3-4 times higher than InteractE when degree is 1 or 2.

\subsection{Performance on relation types}

We compare DeepE with some baseline methods on different relation types. Overall, DeepE achieves the best results on the majority of metrics. DeepE's advantage is particularly prominent on more difficult relations, e.g. head prediction of 1-N and N-N; tail prediction of N-1. The results validate that DeepE is robust on difficult relations.

\section{Related works}
ConvE~\cite{dettmers2018convolutional} is a pioneer work to use convolutional neural networks for KGE. Convolution increases the interactions between the head entity and the relation while keeping parameter efficiency. Later, InteractE~\cite{vashishth2020interacte}, ArcE~\cite{ren2020knowledge} and JointE~\cite{zhou2022jointe} explored the idea further. More interactions are obtained by adding filters, permutating features and changing convolution functions. Compared to these methods, DeepE can learn better on shallow features.

ArcE~\cite{ren2020knowledge} also applies identity mapping by using a skip connection from the input layer to the output layer. Their learning function can be viewed as an addition of a linear function and a non-linear function, while DeepE contains functions with more levels of non-linear orders.

The idea of DeepE's project network is similar with the ones in TransH~\cite{wang2014knowledge} and TransR~\cite{lin2015learning}. The difference is the project function they used is linear, while the one DeepE used is ResNet. The reason is that feature extraction function of DeepE is also non-linear, so a linear function may not suffice for the projection.

Some recent graph based methods~\cite{dai2022mrgat,vashishth2019composition,xie2020reinceptione,wu2021disenkgat} used neighborhood information for knowledge graph completion, while traditional KGE methods only focus on scoring a single triple. The graph based methods often employ KGE methods as feature extraction kernels~\cite{dai2022mrgat,wu2021disenkgat}. Generally, the more powerful the kernel, the better results a graph baseed method may achieve. Hence, the advance of DeepE may further improve the performance of graph based methods.

\section{Conclusion}
In the paper, we propose a deep neural network for knowledge graph embedding. DeepE outperforms the SOTA methods and shows outstanding advantages on robustness. The experiments validate our hypothesis: shallow functions suffice for learning on simple relations, and deep functions serve more for learning on complex relations. Unlike other neural network based KGE methods, DeepE no longer suffers from the performance degradation problem on simple relations, and hence can be very deep while still gaining performance improvement.

For future work, we will explore other learning kernels rather than simple MLP. For example, CNN is parameter effective and has been proved to be useful in many KGE methods. Another research direction is to extend the idea of DeepE to other graph-based learning problem, such as graph neural networks, since the hybrid learning functions seem to be suitable for the graph data.

\bibliography{Deepkgc}

\end{document}